\documentclass[journal]{IEEEtran}
\usepackage{cite}
\usepackage{placeins}
\usepackage{CJKutf8}
\usepackage{float}
\usepackage{amsmath,amssymb,amsfonts}
\usepackage{algorithmic}
\usepackage{graphicx}
\usepackage{textcomp}
\usepackage{multirow}
\usepackage{caption}
\usepackage{placeins}
\usepackage{url}
\usepackage{hyperref}
\usepackage{tabularx}
\usepackage{multirow}
\usepackage{array}
\usepackage{booktabs}
\usepackage{hhline}
\usepackage{makecell}
\usepackage{CJKutf8}
\hypersetup{
colorlinks=true,
linkcolor=black,
citecolor=black
}
\begin{document}
\captionsetup{justification=centering}

\title{Comparative Analysis of Pre-trained Deep Learning Models and DINOv2 for Cushing’s Syndrome Diagnosis in Facial
Analysis}
\author{
    Hongjun Liu\textsuperscript{†}, 
    Changwei Song\textsuperscript{†}, 
    Jiaqi Qiang\textsuperscript{†}, 
    Jianqiang Li\textsuperscript{‡}, 
    Hui Pan\textsuperscript{‡} ,
    Lin Lu\textsuperscript{‡} ,
    Xiao Long\textsuperscript{‡} ,
    Qing Zhao\textsuperscript{*} ,
   Jiuzuo Huang\textsuperscript{*} ,
    Shi Chen\textsuperscript{*}

    \thanks{Hongjun Liu, Changwei Song, and Jiaqi Qiang contributed equally to this work.}
    \thanks{Hongjun Liu, Changwei Song, Jianqiang Li, and Qing Zhao are with School of Software Engineering, Beijing University of Technology, Beijing, China.}
    \thanks{Jiaqi Qiang, Lin Lu, Hui Pan, and Shi Chen are with Key Laboratory of Endocrinology of National Health Commission, Department of Endocrinology, Peking Union Medical College Hospital, Chinese Academy of Medical Sciences and Peking Union Medical College, Beijing 100730, China.}
    \thanks{Xiao Long and Jiuzuo Huang are with Department of Plastic Surgery, Peking Union Medical College Hospital, Chinese Academy of Medical Sciences and Peking Union Medical College, Beijing 100730, China.}
    \thanks{Hui Pan is with State Key Laboratory of Complex Severe and Rare Diseases, Peking Union Medical College Hospital, Chinese Academy of Medical Sciences and Peking Union Medical College, Beijing 100730, China.}
    \thanks{Qing Zhao, Jiuzuo Huang, and Shi Chen are the corresponding authors (Emails: qing.zhao@example.com, hjz1983@126.com, cs0083@126.com).}
    \thanks{This work was supported by National key research \& development plan of China, major project of prevention and treatment for common diseases (2022YFC2505300, subproject: 2022YFC2505304).}
}

\maketitle
\begin{abstract}
Cushing's syndrome is a clinical condition caused by excessive secretion of glucocorticoids from the adrenal cortex. 
It typically manifests with symptoms such as the moon facies and plethora, making facial data a crucial diagnostic criterion. 
Recent studies have employed pre-trained convolutional neural networks for the automatic diagnosis of Cushing's syndrome using frontal facial image data.
However, Cushing's syndrome often presents global facial features, while convolutional neural networks are more efficient in tasks with prominent local features, resulting in a weaker capacity to capture global characteristics.
Transformer-based visual models, such as ViT and SWIN, offer innovative approaches to image classification. 
These models effectively capture long-range dependencies and global features in the input data through self-attention mechanisms. 
Recently, the foundational model DINOv2, which also utilizes the visual Transformer architecture, has garnered significant interest.
The performance of Transformer-based visual models and foundation vision models in diagnosing clinical facial data for Cushing's syndrome remains to be validated.
In this study, we compared the performance of various pre-trained deep learning models, including convolutional neural networks, Transformer-based vision models, and foundational vision models such as DINOv2, to evaluate their effectiveness in transfer learning environments. Additionally, we analyze the gender bias in the models and examine the impact of freezing mechanisms on the performance of the DINOv2 model.
Our findings indicate that in our clinical dataset of Cushing's syndrome, both Transformer-based visual models and the DINOv2 visual foundation model outperformed convolutional neural networks, with the ViT model achieving the highest F1 score of 85.74\%. In terms of the model's gender bias, we found that both the pre-trained model and DINOv2 had significantly higher accuracy for female samples compared to male samples.
Additionally, the DINOv2 model showed significantly improved performance when employing a freezing mechanism compared to not using it. 
In conclusion, Transformer-based visual models and DINOv2 demonstrate strong feasibility for classification tasks in Cushing's syndrome.
The source code is publicly available at: \url{https://github.com/songchangwei/Cushing-Disease-Diagnosis}.

\end{abstract}
\IEEEpeerreviewmaketitle

\section{Introduction}
\IEEEPARstart 
{C}{ushing's} syndrome (CS) is a rare and potentially life-threatening endocrine disorder characterized by chronic excess cortisol. 
This condition leads to various system alterations and is often associated with severe metabolic complications, neuropsychiatric disorders, and characteristic morphological changes, such as central obesity, supraclavicular fat accumulation, skin thinning, purple striae, proximal muscle weakness, fatigue, hypertension, glucose intolerance, acne, hirsutism, and neurological deficits~\cite{arnaldi2003diagnosis}. 
These features pose significant challenges for clinical diagnosis and impact the effectiveness of treatment.
Early diagnosis and treatment can normalize the increased morbidity and mortality associated with Cushing's syndrome~\cite{berr2017persistence}. 
However, the current diagnostic process typically takes 2 to 6 years, highlighting an urgent need for new screening and diagnostic methods.~\cite{psaras2011demographic}.
In recent years, the rapid advancement of artificial intelligence has led to the widespread application of deep learning models in disease diagnosis. 
These models not only enhance diagnostic accuracy but also accelerate the decision-making process for clinicians. 
With increasing data volumes and improved computational capabilities, deep learning shows significant potential in disease prediction and personalized medicine.
Despite the importance of classification models in clinical diagnosis, research specifically focused on the classification of Cushing's syndrome remains limited.

Patients with Cushing's syndrome typically exhibit distinct facial features, such as moon facies, plethora, acne, and hirsutism. 
These characteristics serve as critical indicators for preliminary screening. 
Utilizing facial image data combined with deep learning techniques can facilitate an effective initial diagnosis of Cushing's syndrome.
Currently, some studies have utilized facial imaging and artificial intelligence technologies for the diagnosis of Cushing's syndrome.
Popp et al.~\cite{popp2019computer} analyzed 82 patients with Cushing's syndrome and 98 controls. 
For each participant, standard frontal and lateral photographs were taken. 
Using manual methods and the Facial Image Diagnostic Aid software, 52 key points from the frontal images and 36 from the lateral images were extracted. 
Geometric features were derived from the key point locations, while texture features were obtained using Gabor wavelet similarity functions. 
A maximum likelihood classifier was employed to categorize these features. 
The classification results indicated an overall accuracy of 10/22 (45.5\%) for male patients, 26/32 (81.3\%) for male controls, 34/60 (56.7\%) for female patients, and 43/66 (65.2\%) for female controls.
The aforementioned model's limitation is its reliance on manual annotation. 
Therefore, Wei et al.~\cite{wei2020deep} proposed a fully automated diagnostic method for Cushing's syndrome and acromegaly based on facial images. 
They employed dlib~\cite{king2009dlib} for facial alignment and utilized its pre-trained landmark detection model for face detection. 
After extracting the face, non-facial pixels were masked in black to minimize confusion.
To retain eyebrow features, crucial for acromegaly diagnosis, the upper contour of facial landmarks was elevated to 1/7 of the total facial height. 
The model was fine-tuned using a convolutional neural network pre-trained on ImageNet~\cite{deng2009imagenet} (specifically the VGG~\cite{simonyan2014very} model). 
Key regions associated with the diagnosis of Cushing's syndrome were identified through local occlusion testing on the fine-tuned model.
Experimental results demonstrated an area under the receiver operating characteristic curve (AUC) of 0.9647 for Cushing's syndrome and 0.9556 for acromegaly, with accuracies of 0.9593 and 0.9479, respectively, and recall rates of 0.7593 and 0.8089.
The experimental results of this study indicate that AI algorithms can effectively identify Cushing's syndrome based solely on facial images of patients, without the need for pre-existing medical knowledge.
However, Cushing's syndrome typically presents with global facial features, while convolutional neural networks (CNNs) are more effective for tasks with distinct local features, resulting in a diminished capacity for capturing global characteristics. 
Consequently, CNNs may struggle to effectively identify additional facial features relevant to the diagnosis of Cushing's syndrome.

The impressive success of Transformer~\cite{vaswani2017attention} models in natural language processing, as seen with BERT~\cite{devlin2018bert}, RoBERTa~\cite{liu2019roberta}, and GPT~\cite{radford2018improving,radford2019language,brown2020language}, has inspired their application in computer vision, resulting in models like ViT~\cite{dosovitskiy2020image} and SWIN~\cite{liu2021swin}. 
Unlike convolutional neural networks, Transformer-based vision models can capture global features rather than just local convolutions, offering a distinct advantage in handling complex global dependencies. 
Moreover, these models leverage large-scale pre-training and self-supervised learning to acquire rich representations from vast amounts of unlabeled data.
In recent years, foundational models (FM) in computer vision, such as DINOv2~\cite{oquab2023dinov2}, SAM~\cite{kirillov2023segment} and SAM2~\cite{ravi2024sam2} have gained significant attention. 
Compared to pre-trained models, foundational models utilize larger training datasets, demonstrating enhanced robustness and generalization across various applications, including image classification, retrieval, depth estimation, and semantic segmentation.
The development of foundational models in computer vision has prompted many medical researchers to explore their applications in medical image analysis.
For instance, MedSAM~\cite{MedSAM} is a refined version of the SAM model tailored for medical data, enabling image segmentation of various diseases in medical imaging.
Huix et al.~\cite{huix2024natural} explored the application of DINOv2 in medical image classification and found it to outperform ImageNet-pretrained models in effectiveness. 
However, the internal mechanisms of these models in the clinical diagnosis of Cushing's syndrome based on facial data require further investigation. 
Specifically, a comparative analysis is needed to determine if the foundational models consistently surpass pre-trained deep learning models in diagnosing Cushing's syndrome.

\begin{figure*}[!ht] 
    \centering\includegraphics[width=1\linewidth]{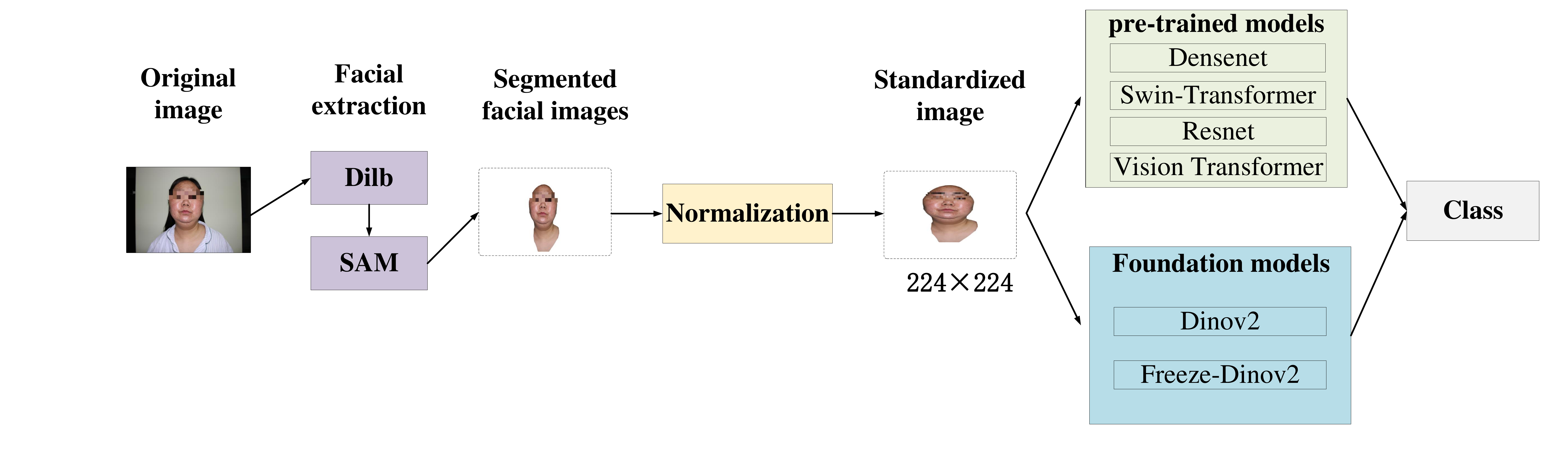}
    \caption{Flow diagram of this study. }
    \label{fig:suicide} 
\end{figure*}

In this study, we compared the diagnostic performance of pre-trained models, including Densenet~\cite{huang2017densely}, ResNet~\cite{he2016deep}, ViT~\cite{alexey2020image}, and Swin Transformer~\cite{liu2021swin}, against Dinov2 (a visual backbone model) using clinical data from Cushing's syndrome.
Specifically, we investigated the impact of the freezing mechanism on performance by fixing the Dinov2 backbone while fine-tuning other layers. 
Our dataset comprised 2D frontal facial images from 343 patients, including 49 from the disease group and 294 from the control group. The process of this study is shown in Figure~\ref{fig:suicide}.
Based on the experimental results, we found that the freezing approach using Dinov2 consistently outperformed training without parameter freezing. 
Furthermore, the attention-based vision model ViT surpassed both convolutional neural networks and the visual backbone model Dinov2, achieving an optimal F1 score of 85.74\%. 
Lastly, we observed a significant bias in the model’s performance regarding gender, with diagnostic accuracy for female Cushing's syndrome consistently exceeding that for males.

\section{Related work} \label{sec:related} 

\subsection{Foundational Models for Image Classification}
In computer vision, foundational models are increasingly applied to various tasks.
For example, the Segment Anything Model (SAM)~\cite{kirillov2023segment} is a foundational model that performs image segmentation using text, boxes, and points as prompts.It has been widely applied in medical image segmentation tasks ~\cite{wu2023medical,chen2024ma,cheng2023sam,zhang2023input}.
Liu et al.~\cite{liu2024vision} introduced VISION-MAE, another foundational model that supports both image segmentation and classification.
This model is based on a Swin Transformer~\cite{liu2021swin} masked autoencoder and was trained on a dataset of 2.5 million clinical images, including CT, MRI, PET/CT, radiography (X-ray), and ultrasound.
Another representative model is DINOv2~\cite{oquab2023dinov2}, an unlabeled self-distillation method utilizing the Vision Transformer~\cite{dosovitskiy2020image} architecture, renowned for its superior performance compared to other self-supervised and supervised learning methods.

Vision foundation models, exemplified by DINOv2, have been extensively researched and applied in medical imaging.
Baharoon et al.~\cite{baharoon2023towards} employed the DINOv2 model for radiological image analysis, conducting over 100 experiments across various modalities (X-ray, CT, and MRI) for tasks including disease classification and organ segmentation. 
The results demonstrated that DINOv2 exhibited strong cross-task generalization capabilities compared to supervised, self-supervised, and weakly supervised models.
Veasey et al.~\cite{veasey2024parameter} applied low-rank adaptation (LoRA)~\cite{hu2021lora} to the foundation model DINOv2 for image classification in the assessment of malignant lung tumors.
Huix et al.~\cite{huix2024natural} evaluated the performance of five foundation models—SAM, SEEM~\cite{zou2024segment}, DINOv2, BLIP~\cite{li2022blip}, and CLIP~\cite{radford2021learning}—across four comprehensive medical imaging datasets. 
When paired with classifiers, DINOv2 outperformed the baseline, demonstrating strong applicability in medical imaging.

These studies underscore the potential and effectiveness of the DINOv2 model, emphasizing its significant role in medical image analysis.
However, it is important to note that most of these studies rely on public datasets.
This reliance on public data results in a lack of validation with clinical data, which is crucial for a comprehensive assessment and real-world applicability in medical settings.
\subsection{Artificial Intelligence for Cushing's Syndrome}
The rapid advancement of artificial intelligence has prompted researchers to apply it to the study of Cushing's syndrome.
For instance, Zhang et al.~\cite{zhang2021electronic} developed a machine learning model to predict immediate relief after transsphenoidal surgery for Cushing's syndrome using structured features and free-text data from electronic medical records (EMR).

They employed four machine learning algorithms: Multilayer Perceptron
 (MLP)~\cite{rumelhart1986learning}, Support Vector Machine (SVM)~\cite{cortes1995support}, Random Forest (RF)~\cite{breiman2001random}, and Logistic Regression (LR)~\cite{cox1958regression}, achieving maximum AUC values of 0.759, 0.733, 0.678, and 0.699, respectively.
For example, Fan et al.~\cite{fan2021toward} explored a neural network approach based on factorization machines to predict the recurrence of Cushing's syndrome.
The study utilized 17 clinical features, including age, BMI, disease duration, and tumor size, from a dataset comprising 354 patients with postoperative initial remission and long-term follow-up data. 
Compared to traditional machine learning algorithms (GBDT~\cite{friedman2001greedy}, AdaBoost~\cite{freund1997decision}, XGBoost~\cite{chen2016xgboost}, etc.), their method achieved the highest AUC of 0.869 and the lowest log loss of 0.256.\\
\indent While many studies have applied AI techniques to various tasks related to Cushing's syndrome, this research primarily focuses on its diagnosis and classification.
For example, Isci et al.~\cite{isci2021machine} developed machine learning models for classifying Cushing's syndrome based on clinical features, laboratory test results, and imaging data. 
The dataset included 241 subjects (183 females and 58 males, mean age ± SD = 52.02 ± 13.33 years). The machine learning models included Support Vector Machine (SVM), K-Nearest Neighbors (KNN)~\cite{altman1992introduction}, Logistic Regression (LR), Linear Discriminant Analysis (LDA)~\cite{tharwat2016principal}, Decision Trees (DT) based on Classification and Regression Trees (CART)~\cite{loh2011classification}, Random Forest (RF), AdaBoost, and Gradient Boosting (GB). 
The results indicated that the RF algorithm outperformed others, achieving an F1 score of 89.5\% and an accuracy of 91.4\%.
However, the data used in this study includes laboratory test results and medical imaging, which are often difficult to obtain. 
In contrast, facial image data from patients is more readily accessible. 
Consequently, some studies have opted to use facial images for diagnosing Cushing's syndrome.
Popp et al.~\cite{popp2019computer} analyzed 82 patients with Cushing's syndrome and 98 controls. 
For each participant, standard frontal and lateral photographs were taken. 
Using manual methods and the Facial Image Diagnostic Aid software, 52 key points from the frontal images and 36 from the lateral images were extracted. 
Geometric features were derived from the key point locations, while texture features were obtained using Gabor wavelet similarity functions. 
A maximum likelihood classifier was employed to categorize these features. 
The classification results indicated an overall accuracy of 10/22 (45.5\%) for male patients, 26/32 (81.3\%) for male controls, 34/60 (56.7\%) for female patients, and 43/66 (65.2\%) for female controls.
Kosilek et al.~\cite{kosilek2013automatic} employed a similar approach but utilized the function P (including the scalar product of phases) as texture features and L (length of edge differences) as geometric features.
These two studies validated the feasibility of using facial images for diagnosing Cushing's syndrome. 
However, their limitation lies in the reliance on manual methods, lacking full automation in diagnosis.
Wei et al.~\cite{wei2020deep} proposed a fully automated diagnostic method for Cushing's syndrome and acromegaly based on facial images. 
They utilized dlib~\cite{king2009dlib} for face alignment and employed a pre-trained keypoint detection model for facial detection. 

Experimental results demonstrated an area under the receiver operating characteristic curve (AUC) of 0.9647 for Cushing's syndrome and 0.9556 for acromegaly, with accuracies of 0.9593 and 0.9479, respectively, and recall rates of 0.7593 and 0.8089.\\
\indent These studies highlight the potential and applicability of deep learning techniques in the diagnosis of Cushing's syndrome based on facial images.

However, Cushing's syndrome often presents global features in facial images, while convolutional neural networks (CNNs) are more efficient at processing tasks with prominent local features. 
This limits their ability to capture global characteristics, potentially hindering the effective identification of additional facial features relevant to the diagnosis of Cushing's syndrome.
Due to the significant success of Transformer models in natural language processing, such as BERT, RoBERTa, and GPT, researchers have introduced them into computer vision, developing models like ViT and SWIN. 
Compared to convolutional neural networks, Transformer-based visual models more effectively capture global features in images rather than relying solely on local convolutional operations. 
This capability shows marked advantages in addressing complex global dependencies. 
Additionally, Transformer models can learn rich representations from large-scale pre-training and self-supervised learning using vast amounts of unlabeled data.
In recent years, foundational models (FMs) have gained significant attention in computer vision, exemplified by DINOv2~\cite{oquab2023dinov2}, SAM~\cite{kirillov2023segment}, and SAM2~\cite{ravi2024sam2}. 
Compared to traditional pre-trained models, foundational models leverage larger training datasets, demonstrating enhanced robustness and generalization across various applications, including image classification, retrieval, depth estimation, and semantic segmentation.
However, the internal mechanisms of these models in the clinical diagnosis of Cushing's syndrome based on facial data require further investigation. 
Specifically, comparative analyses are needed to determine whether foundational models consistently outperform pre-trained deep learning models in diagnosing Cushing's syndrome.

\section{Methods}

In this study, we conducted a comparative analysis of a clinical dataset of facial images from patients with Cushing's syndrome using ImageNet pre-trained deep learning models and baseline models. 
This research highlights the impact of the freezing mechanism of the DINOv2 model on performance outcomes. 
Furthermore, we specifically examined the potential biases in these models when diagnosing patients of different sexes within the clinical dataset, with comprehensive results presented in the subsequent section.

\subsection{Transfer learning on ImageNet pre-trained models}

In this study, we compared several popular pre-trained models, focusing on their parameter counts and performance variations. The models evaluated include Densenet~\cite{huang2017densely}, ResNet~\cite{he2016deep}, Swin Transformer~\cite{liu2021swin}, and Vision Transformer (ViT)~\cite{alexey2020image}.
These models demonstrated superior performance in large-scale image recognition tasks.

\begin{itemize}
    \item \textbf{DenseNet~\cite{huang2017densely}:} DenseNet, or Densely Connected Convolutional Networks is a deep learning model proposed by Gao Huang et al. in 2017. It improves information flow and gradient propagation by creating dense connections, where each layer is directly connected to all subsequent layers. Such connections maximize feature reuse and alleviate the vanishing gradient problem, enhancing overall performance. DenseNet comprises Dense Blocks and Transition Layers. Within a Dense Block, each layer is connected to all preceding layers, allowing information to be merged through feature concatenation. Transition Layers include convolutional and pooling layers to control the number of feature maps and adjust their dimensions. Key parameters are growth rate (the number of feature maps added at each layer), bottleneck layers ($1 \times 1$ convolutions to reduce computational load), and compression rate (the factor by which feature maps are reduced in Transition Layers). By leveraging efficient feature reuse and improved gradient flow, DenseNet achieves remarkable performance with fewer parameters, making it highly suitable for tasks such as image classification, object detection, and semantic segmentation. It has demonstrated exceptional results in applications such as medical imaging, autonomous driving, and security monitoring. In this study, we employ several pre-trained DenseNet models: DenseNet121, DenseNet161, DenseNet169, and DenseNet201.

    \item \textbf{ResNet~\cite{he2016deep}:} ResNet, or Residual Network, is a deep learning model introduced by Kaiming He et al. in 2015, designed to address the degradation problem encountered during the training of deep neural networks. Its core innovation is the introduction of residual blocks, which utilize skip connections to add the input directly to the output, enabling each block to learn the residual between the input and output rather than fitting a complex mapping function. A standard residual block consists of two convolutional layers, each followed by batch normalization and a ReLU activation function, with the input added directly to the output of the final convolution. This structure simplifies gradient backpropagation and mitigates issues related to vanishing and exploding gradients in deep networks. As a result, ResNet enables the effective training of deeper networks, significantly enhancing performance, with variants like ResNet-50, ResNet-101, and ResNet-152 achieving remarkable results on large datasets such as ImageNet. The residual architecture facilitates efficient information transfer and smooth gradient propagation, making ResNet highly applicable to tasks like image recognition, object detection, and semantic segmentation. It has become a foundational framework in modern deep learning model design. The pre-trained models selected for this study are ResNet-18, ResNet-34, ResNet-50, ResNet-101, and ResNet-152.

    \item \textbf{ViT~\cite{alexey2020image}:} ViT, or Vision Transformer, is a novel computer vision model introduced by Google Research in 2020 that incorporates the Transformer architecture into the image processing, challenging the dominance of traditional convolutional neural networks (CNNs). The central idea of ViT is to divide images into fixed-size patches, which are then flattened and linearly embedded into a high-dimensional vector space, resulting in a series of patch embeddings. These embeddings, along with position encodings, serve as inputs to the Transformer encoder, which processes them through self-attention mechanisms and feed-forward networks. This architecture allows the model to capture long-range dependencies across the entire image, unlike CNNs, which are restricted to local receptive fields. As a result, ViT proves to be more effective for handling complex visual tasks. When pretrained on large datasets like ImageNet, ViT can achieve or exceed the performance of state-of-the-art convolutional networks with fewer computational resources. Its design has sparked a paradigm shift in computer vision, showcasing the potential of Transformers in image data processing and demonstrating outstanding performance in tasks such as image classification, object detection, and semantic segmentation. The success of ViT has also led to further research into adapting the Transformer architecture for additional visual tasks and multimodal learning scenarios. In this study, we examine the following pre-trained models: ViT-B16, ViT-B32, ViT-L16, and ViT-L32.

    \item \textbf{Swin Transformer~\cite{liu2021swin}:} Swin Transformer, or Hierarchical Vision Transformer, is a visual Transformer model introduced by the Microsoft Research Asia team in 2021, designed to address the computational and memory bottlenecks of traditional Vision Transformers (ViT) when processing high-resolution images. The primary innovation of the Swin Transformer lies in its hierarchical structure and shifted window attention mechanism. The model first divides the image into non-overlapping fixed-size patches, performing local self-attention computations within each patch, which significantly reduces computational complexity. To capture global information across patches, it employs a sliding window strategy that shifts the window position between adjacent layers, facilitating the gradual integration of cross-patch information. Swin Transformer consists of multiple stages, with each stage further divided into several Swin Transformer blocks. As the depth of the network increases, the patch size enlarges while the feature map resolution decreases, leading to hierarchical feature representations. This design effectively preserves global image information while substantially enhancing computational efficiency and scalability. Swin Transformer has demonstrated outstanding performance across various computer vision tasks, such as image classification, object detection, and semantic segmentation, particularly excelling in high-resolution image processing and scene understanding tasks. Its efficient hierarchical attention mechanism provides new insights and directions for optimizing visual Transformer models. In this study, we evaluate several pre-trained Swin Transformer models, including Swin-T, Swin-S, Swin-B, and Swin-L.
    \end{itemize}

\subsection{DINOv2: vision transformer-based computer vision foundation model}
DINOv2, or Self-Distillation with No Labels v2, is a novel self-supervised learning algorithm introduced by the Meta AI Research team in 2023 to enhance visual representation learning. 
Building on the original DINO, DINOv2 incorporates several improvements, including more sophisticated data augmentation strategies and a robust teacher-student model architecture, resulting in higher-quality unsupervised feature representations.
Specifically, DINOv2 utilizes Vision Transformer (ViT) as its backbone, employing the principle of self-supervised knowledge distillation. 
It trains a teacher model to generate pseudo-labels, which guide the learning of a student model. 
Unlike traditional self-supervised methods, DINOv2 introduces a dynamic knowledge distillation strategy that continuously updates the teacher model during training, enabling the student model to learn more refined and diverse features.
Additionally, DINOv2 integrates advanced techniques such as multi-scale feature extraction and global/local contrastive learning, demonstrating remarkable performance across various visual tasks, including image classification, object detection, and semantic segmentation. 
Importantly, DINOv2 can effectively train on large-scale unlabeled datasets without requiring extensive labeled data, significantly enhancing the model's generalization capabilities and applicability. 
This positions DINOv2 as a significant breakthrough in the field of self-supervised learning, advancing unsupervised visual representation learning technologies.
We investigate various DINOv2 models: DINOv2-s, DINOv2-b, and DINOv2-l. 
This study emphasizes the impact of the freezing mechanism on DINOv2's performance, fixing the backbone while fine-tuning other layers for comparative analysis.

\section{Experiments} \label{sec:experiments}
\subsection{Dataset}

\begin{table}[h!]
\centering
\caption{Gender distribution in control and disease groups across different sets}
\label{tab:datasets}
\begin{tabular}{|c|c|c|c|}
\hline
& \textbf{gender} & \textbf{Control Group} & \textbf{Disease Group} \\
\hline
\multirow{2}{*}{\textbf{Training Set}} & Male & 13 & 3 \\
& Female & 85 & 14 \\
\hline
\multirow{2}{*}{\textbf{Validation Set}} & Male & 16 & 4 \\
& Female & 82 & 12 \\
\hline
\multirow{2}{*}{\textbf{Test Set}} & Male & 11 & 4 \\
& Female & 87 & 12 \\
\hline
\end{tabular}
\label{table:gender_distribution}
\end{table}
In this study, we analyzed 2D frontal facial images of 343 patients with Cushing's syndrome.
\begin{table*}[!ht]
\centering
\caption{Performance of DinoV2 and pre-trained models (ViT,Swin,Densenet,ResNet) in the diagnosis of Cushing's syndrome.}
\label{tab_fwsc}
\begin{tabular}{|c|c|c|c|c|c|} 
\hline
Model          &Model type          & Accuracy             & precision      &  recall & F1 \\ 

\hline
\multirow{6}{*}{Dinov2}& Dinov2-s& 0.9474& 0.8571& 0.7500& 0.8000\\
    &Dinov2-b& 0.9474& 0.7778& 0.8750& 0.8235\\
    &Dinov2-l& 0.9474& 0.8571& 0.7500& 0.8000\\
    &Dinov2-s-unfrez& 0.8772& 0.5455& 0.7500& 0.6316\\
    &Dinov2-b-unfrez& 0.9386& 0.8000& 0.7500& 0.7742\\
    &Dinov2-l-unfrez& 0.8684& 0.5217& 0.7500& 0.6154\\
\hline
\multirow{4}{*}{VIT}    & Vit\_B\_16  & 0.9561 & 1.0000 & 0.6875 & 0.8148\\
    &Vit\_B\_32 &0.9649 &1.0000&0.7500&0.8571\\
    &Vit\_L\_16 &0.9561 &0.8667&0.8125&0.8387\\
    &Vit\_L\_32 &0.9386  &0.9091&0.6250 &0.7407\\
\hline
\multirow{4}{*}{Swin}& Swin-T& 0.8333& 0.3846& 0.3125& 0.3448\\
    &Swin-S& 0.6579& 0.2909& 1.0000& 0.4507\\
    &Swin-B& 0.9298& 0.7222& 0.8125& 0.7647\\
    &Swin-L& 0.8947& 1.0000& 0.2500& 0.4000\\
\hline
\multirow{4}{*}{Densenet}   &Densenet121& 0.8860& 0.7143& 0.3125& 0.4348\\
    &Densenet161& 0.8772& 0.5333& 1.0000& 0.6957\\
    &Densenet169& 0.8947& 0.5909& 0.8125& 0.6842\\
    &Densenet201& 0.8947& 0.5833& 0.8750& 0.7000\\
    
\hline
\multirow{5}{*}{Resnet}  & Resnet18& 0.8684& 0.5200 &0.8125& 0.6341\\
    &Resnet34&0.8860&0.5517&1.0000&0.7111\\
    &Resnet50& 0.8860& 0.5556& 0.9375& 0.6977\\
    &Resnet101& 0.8860& 0.6154& 0.5000& 0.5517\\
    &Resnet152& 0.9035& 0.7273& 0.5000& 0.5926\\
\hline
\end{tabular}
\end{table*}
This study was approved by the Ethics Committee of Peking Union Medical College Hospital.
Our dataset comprises 343 2D images, including 49 samples from the disease group and 294 from the control group. 
Each group contains one frontal view. 
Patients with Cushing's syndrome were recruited from the Endocrinology Department at Peking Union Medical College Hospital between March 2023 and April 2024. 
Inclusion criteria included: aged 18 to 65 years, regardless of gender; confirmed diagnosis via surgical pathology or clinical diagnosis by an endocrinologist based on clinical presentation, hormone levels, and diagnostic tests; and within 10 days before or after surgery.  
Exclusion criteria included: patients without a clear diagnosis per clinical guidelines; poor quality facial images; history of significant facial surgery, trauma, cosmetic procedures (e.g., hyaluronic acid injections), or orthodontics; and those with systemic diseases such as scleroderma, systemic lupus erythematosus, or dermatomyositis that may alter facial appearance.
\begin{figure*}[!ht] 
    \centering\includegraphics[width=1\linewidth]{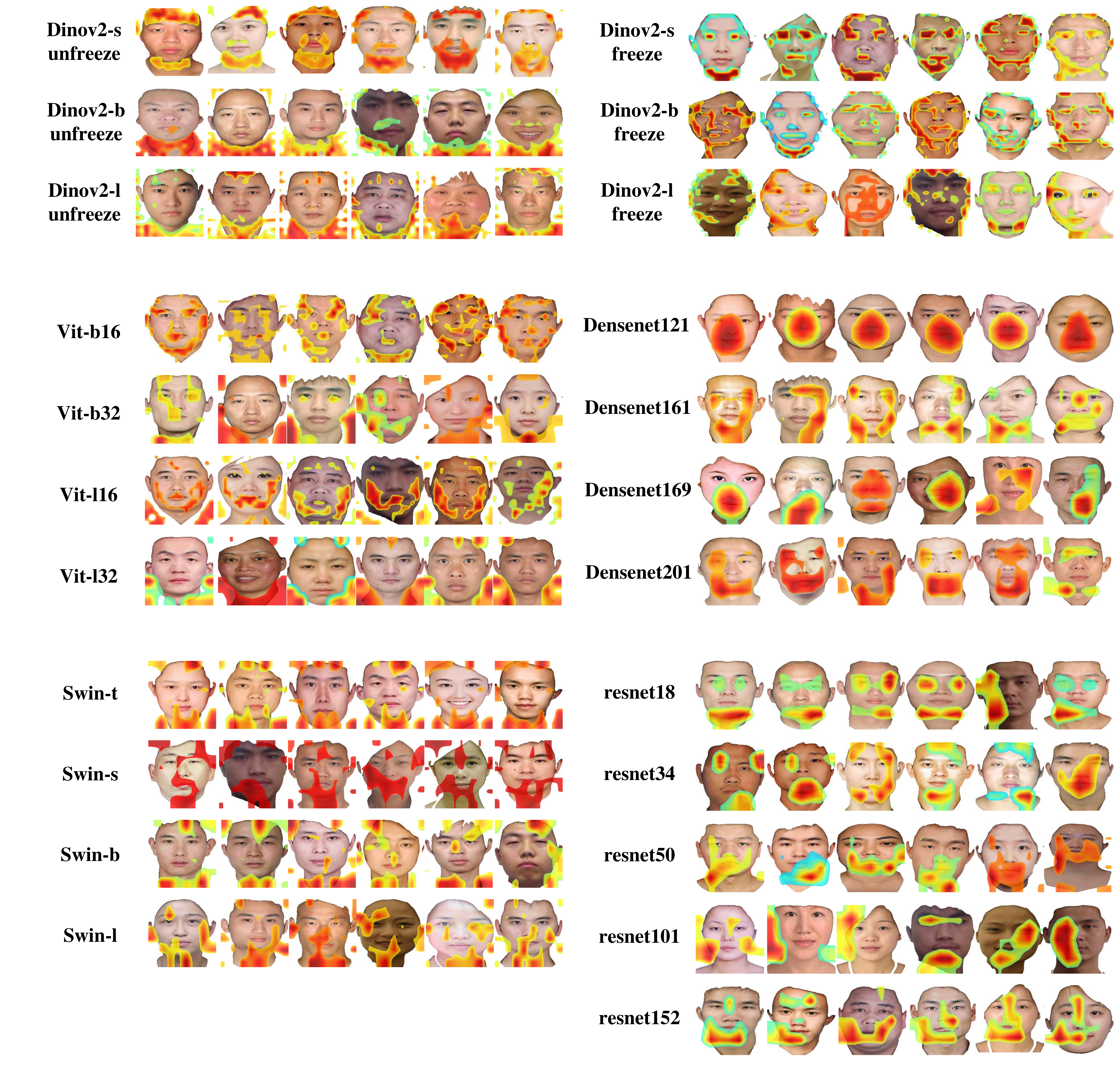}
    \caption{Activation maps of the pre-trained model and the foundational model. }
    \label{fig:suicide1} 
\end{figure*}
The control group for Cushing's syndrome was recruited from the Endocrinology and Plastic Surgery Departments at Peking Union Medical College Hospital between April 2022 and April 2024. 
Inclusion criteria included: aged 18 to 65 years, regardless of gender; diagnosed with obesity ($BMI \ge  28 \text{kg/m}^2 $) or overweight ($ 24 \text{kg/m}^2 \le  BMI < 28 \text{kg/m}^2 $), or matched BMI to Cushing's syndrome patients; and no typical features of Cushing's syndrome, such as moon facies, plethora, upper lip hair, acne, buffalo hump, supraclavicular fat pads, slender limbs, or abdominal striae. 
Exclusion criteria included: poor quality facial images; history of exogenous corticosteroid use; significant facial surgery, trauma, cosmetic procedures (e.g., hyaluronic acid injections), or orthodontics; and systemic diseases such as scleroderma, systemic lupus erythematosus, or dermatomyositis affecting facial skin, which could alter appearance.
We randomly divided the samples into training, testing and validation sets in a 1:1:1 ratio, with the test set used solely for model evaluation. 
More details of the dataset are shown in Table~\ref{tab:datasets}.
\subsection{Implementation details}
In this study, we employed various metrics to compare model performance, including accuracy, precision, and recall. 
To address potential data imbalance, we also calculated the F1 score for a comprehensive evaluation. 
Additionally, we extracted activation maps to assess feature extraction capabilities.
All experiments were conducted using an NVIDIA GeForce RTX 4090 equipped with 24GB of GPU memory.
Our model was trained for 200 epochs with a batch size of 32, utilizing the Stochastic Gradient Descent (SGD)~\cite{bottou2010large} optimizer and a learning rate of $5 \times 10^{-4}$. 
Throughout the training, we used the cross-entropy loss function. 
All implementations were based on the PyTorch~\cite{paszke2019pytorch} framework.

\subsection{Face Extraction}\label{sec:methods:sub1}
Due to variations in photographers and shooting environments, background differences in images can affect experimental results. 
To mitigate this interference, we extract facial regions from the images. 
Initially, we employ the Dlib library, a powerful tool widely used in computer vision for tasks such as face extraction and recognition. 
After processing the initial photo with Dlib, most non-facial areas are removed, resulting in an image focused on the face. However, some unwanted elements like hair, clothing, or background may remain. 
Thus, we utilize the SAM model for further facial extraction. 
SAM is a foundational model in computer vision that has demonstrated remarkable performance in segmentation tasks. 
During the extraction process, we place a label point at the center of the face; SAM generates a mask centered on this point, ultimately yielding a clean facial image devoid of any distracting information.

\subsection{Images Standardize and Data Augmentation}
After obtaining the extracted facial images, we standardize them for model input requirements. 
We first remove the white areas surrounding the face to focus the image on the facial features.
Then, we uniformly resample the images to a size of $224  \times 224$, ensuring consistency in dimensions and resolution throughout the analysis.

The dataset comprises 49 cases in the patient group and 294 in the control group, resulting in a class imbalance ratio of 1:6 that may affect model outcomes. 
To address this issue, we implemented data augmentation strategies. 

We increased the number of patient samples in the training set to five times the original, reaching 85 samples, thereby approximating a 1:1 ratio with the control group. 
Our augmentation techniques included horizontal flipping, random center cropping, random brightness adjustments, and random rotation. 
These methods expand the dataset and enhance its diversity, helping to mitigate the impact of data imbalance.
\section{Results}
Our research focuses on the performance of DinoV2 and other pre-trained models (such as ResNet, ViT, DenseNet, and Swin Transformer) in diagnosing Cushing's syndrome using 2D facial data. 
We particularly emphasize the impact of the freezing mechanism on DinoV2's performance. The experimental results are presented in Table~\ref{tab_fwsc}.
\begin{table*}[!ht]
\centering
\caption{Comparative performance of Dinov2 and pre-trained models (ResNet, DenseNet, ViT, Swin) on gender-specific datasets.}
\label{tab_fwsc2}
\begin{tabular}{|c|c|c|c|c|c|c|} 
\hline
Model          &Model type & Model gender           & Accuracy             & precision      &  recall & F1 \\ 
\hline
\multirow{12}{*}{Dinov2}    & \multirow{2}{*}{Dinov2-l}       & Dinov2-l-female     & 0.9596 & 0.9 & 0.75  &0.8182 \\ 
\cline{3-7}
                                 &                              & Dinov2-l-male     & 0.8667 & 0.75 & 0.75 &0.75  \\ 
\cline{2-7}
                                 & \multirow{2}{*}{DINOv2-l-unfrez}    & DINOv2-l-unfrez-female    & 0.9394 & 0.6875 & 0.9167 & 0.7857 \\ 
\cline{3-7}
                                 &                              & DINOv2-l-unfrez-male      & 0.4  & 0.1429& 0.25 & 0.1818\\ 
\cline{2-7}
                                 & \multirow{2}{*}{Dinov2-b} & Dinov2-b-female      & 0.9596& 0.7857 & 0.9167  &0.8462\\ 
\cline{3-7}
                                 &                              & Dinov2-b-male     & 0.8667 & 0.75 & 0.75&0.75  \\ 
\cline{2-7}
                                 & \multirow{2}{*}{DINOv2-b-unfrez}    & DINOv2-b-unfrez-female      & 0.9697 & 0.9091 & 0.8333  &0.8696 \\ 
\cline{3-7}
                                 &                              & DINOv2-b-unfrez-male     & 0.7333 & 0.5 & 0.5 &0.5  \\ 
\cline{2-7}
                                 & \multirow{2}{*}{Dinov2-s}    & Dinov2-s-female     & 0.9596& 0.9 & 0.75  &0.8182 \\ 
\cline{3-7}
                                 &                              & Dinov2-s-male     & 0.8667& 0.75 & 0.75&0.75 \\ 
\cline{2-7}
                                 & \multirow{2}{*}{DINOv2-s-unfrez}    & DINOv2-s-unfrez-female     & 0.9394 & 0.7143 & 0.8333   &0.7692\\ 
\cline{3-7}
                                 &                              & DINOv2-s-unfrez-male     & 0.4667 & 0.25 & 0.5   &0.3333\\ 
\cline{1-7}
\multirow{8}{*}{ViT} & \multirow{2}{*}{Vit-L-16}       & Vit-L-16-female      & 0.9798 & 0.9167 & 0.9167  &0.9167 \\ 
\cline{3-7}
                                 &                              & Vit-L-16-male     & 0.8&0.6667 & 0.5  &0.5714 \\ 
\cline{2-7}
                                 & \multirow{2}{*}{Vit-L-32}    & Vit-L-32-female     & 0.9596& 0.9 & 0.75  &0.8182 \\ 
\cline{3-7}
                                 &                              & Vit-L-32-male     & 0.8& 1 & 0.25&0.4 \\ 
\cline{2-7}
                                 & \multirow{2}{*}{Vit-B-16} & Vit-B-16-female    & 0.9697 & 1& 0.75&0.8571  \\ 
\cline{3-7}
                                 &                              & Vit-B-16-male     & 0.8667 & 1 & 0.5  &0.6667 \\ 
\cline{2-7}
                                 & \multirow{2}{*}{Vit-B-32}    & Vit-B-32-female     &0.9899 & 1 & 0.9167  &0.9565 \\ 
\cline{3-7}
                                 &                              & Vit-B-32-male      & 0.8 & 1& 0.25 &0.4  \\ 

\cline{1-7}
\multirow{8}{*}{Swin}       & \multirow{2}{*}{Swin-T}       & Swin-T-female &0.8586 & 0.375 & 0.25  &0.3 \\ 
\cline{3-7}
                                 &                              & Swin-T-male     & 0.6667 &0.4 & 0.5  &0.4444\\ 
\cline{2-7}
                                 & \multirow{2}{*}{Swin-S}    &Swin-S-female  & 0.6869 & 0.2791 & 1  &0.4364 \\ 
\cline{3-7}
                                 &                              & Swin-S-male      & 0.4667 & 0.3333 & 1  &0.5 \\ 
\cline{2-7}
                                 & \multirow{2}{*}{Swin-B} & Swin-B-female     & 0.9596 & 0.8333 & 0.8333 &0.8333  \\ 
\cline{3-7}
                                 &                              & Swin-B-male     & 0.7333 & 0.5 & 0.75 &0.6 \\ 
\cline{2-7}

                                 & \multirow{2}{*}{Swin-L}    &Swin-L-female      &0.9192 & 1& 0.3333 &0.5 \\ 
\cline{3-7}
                                 &                              & Swin-L-male      & 0.7333 & 0 & 0&0 \\ 
\cline{1-7}
\multirow{8}{*}{Densenet}       & \multirow{2}{*}{Densenet121}       & Densenet121-female &0.9091 & 0.8 & 0.3333 &0.4706 \\ 
\cline{3-7}
                                 &                              & Densenet121-male   & 0.7333 &0.5& 0.25&0.3333\\ 
\cline{2-7}
                                 & \multirow{2}{*}{Densenet161}    &Densenet161-female  & 0.9192 & 0.6 & 1  &0.75 \\ 
\cline{3-7}
                                 &                              & Densenet161-male     & 0.6 & 0.4 & 1  &0.5714 \\ 
\cline{2-7}
                                 & \multirow{2}{*}{Densenet169} & Densenet169-female     & 0.9192 & 0.625 & 0.8333 &0.7143  \\ 
\cline{3-7}
                                 &                              & Densenet169-male    & 0.7333 & 0.5 & 0.75 &0.6 \\ 
\cline{2-7}
                                 & \multirow{2}{*}{Densenet201}    &Densenet201-female    &0.9495 & 0.7333& 0.9167&0.8148\\ 
\cline{3-7}
                                 &                              & Densenet201-male    & 0.5333 & 0.3333 & 0.75 &0.4615 \\ 
\cline{1-7}
\multirow{10}{*}{Resnet}       & \multirow{2}{*}{Resnet18}       & Resnet18-female &0.9293& 0.6471 & 0.9167 &0.7586 \\ 
\cline{3-7}
                                 &                              & Resnet18-male  & 0.4667&0.25& 0.5&0.3333\\ 
\cline{2-7}
                                 & \multirow{2}{*}{Resnet34}    &Resnet34-female & 0.9293 & 0.6316 & 1  &0.7742 \\ 
\cline{3-7}
                                 &                              & Resnet34-male    & 0.6 & 0.4 & 1  &0.5714 \\ 
\cline{2-7}
                                 & \multirow{2}{*}{Resnet50} & Resnet50-female    & 0.9394 & 0.6667 & 1&0.8 \\ 
\cline{3-7}
                                 &                              & Densenet169-male    & 0.7333 & 0.5 & 0.75 &0.6 \\ 
\cline{2-7}
                                 & \multirow{2}{*}{Resnet101}    &Resnet101-female    &0.9091 & 0.6364& 0.5833&0.6087\\ 
\cline{3-7}
                                 &                              & Resnet101-male   & 0.7333 & 0.5 & 0.25&0.3333\\ 
\cline{2-7}
                                 & \multirow{2}{*}{Resnet152}    &Resnet152-female   &0.9495 & 0.8889&0.6667&0.7619\\ 
\cline{3-7}
                                 &                              & Resnet152-male & 0.6 & 0.4 & 1&0.5714\\ 
\cline{1-7}
\end{tabular}
\end{table*}

Furthermore, we compared the performance of the model across different sexes, with the results presented in Table~\ref{tab_fwsc2}.
This section presents key results and discusses the main insights regarding the observed model performance.  \\
\indent Firstly, our study on the classification performance of diagnostic models for Cushing's syndrome indicates that the ViT and DINOv2 models perform the best.
Specifically, the ViT model achieved the highest results with an accuracy of 96.49\% and an F1 score of 85.71\%, while the DINOv2 model yielded the second-best results with an accuracy of 94.74\% and an F1 score of 82.35\%.
In particular, all DINOv2 models frozen by parameter (DINOv2-s, DINOv2-b, DINOv2-l) achieved F1 scores of at least 80\%, while all four ViT models (ViT-B-16, ViT-B-32, ViT-L-16) exceeded an F1 score of 80\%, highlighting their robust architectures.
The optimal DenseNet model achieved an F1 score of 70.00\%, while the best ResNet model reached 71.11\%. 
Among Transformer-based models, the optimal Swin model attained an F1 score of 76.46\%, and both ViT and DINOv2 models exceeded 80\%. 
This demonstrates that Transformer-based visual models outperform convolutional neural networks in diagnosing Cushing's syndrome using facial image data.
To further validate the performance of the models in the diagnosis of Cushing syndrome, we generated the activation maps of the models, as shown in Figure ~\ref{fig:suicide1}, and compared them with medical knowledge. The results show that the activation maps of the ViT model and the DINOv2 model mainly focus on key areas in facial images that align with medical knowledge. This result is highly consistent with the experimental predictions. This consistency indicates that these models can rely on medical knowledge and focus on the correct pathological regions during the classification process.\\
\indent Furthermore,  for the DINOv2 model, without the freezing mechanism, the F1 scores for the three variants—DINOv2-s, DINOv2-b, and DINOv2-l—were 63.16\%, 77.42\%, and 61.54\%, respectively. 
After implementing the freezing mechanism, these scores improved to 80.00\%, 82.35\%, and 80.00\%. 
Specifically, the introduction of the freezing mechanism resulted in increases in the F1 score of 16. 84\%, 4. 93\%, and 18. 46\% for the three models, respectively.
The introduction of the freezing mechanism in DINOv2 not only significantly improved model performance but also accelerated training speed.
This suggests that the features learned through self-supervised learning in DINOv2 exhibit a high generalization capability. 

\section{Discussion}
In this study, we compared the performance of pre-trained models and foundational models in diagnosing Cushing's disease. The results showed that the ViT model and DINOv2 model performed the best. We also generated activation maps of the models to further validate these results. Additionally, we investigated the impact of the freezing mechanism on the DINOv2 model, and the results demonstrated that introducing the freezing mechanism significantly improved the model's performance.

Moreover, we found that larger models do not necessarily outperform smaller ones, whether in base or pre-trained contexts.
For instance, the best-performing model in the ViT series is ViT\_B\_32; in the ResNet series, it is ResNet34; in the Swin Transformer series, the optimal model is Swin-B; and in the DINOv2 series, the top-performing model is DINOv2-b.
This finding suggests that we should choose the most appropriate model rather than the one with the most or fewest parameters.

Finally, when analyzing potential gender bias in models, our results indicate that the DINOv2 model demonstrates significantly higher accuracy for female samples compared to male samples, achieving a maximum F1 score of 75\% for males and 84.62\% for females. 
The pre-trained model exhibits a significantly higher accuracy for female samples compared to male samples.
The discrepancy may arise from a substantial imbalance in the number of male and female samples in the training set. 
To mitigate potential gender bias in model performance, future research should adopt a more comprehensive approach to address this notable gender disparity among patients.

Despite the good performance of pre-trained and foundational models in diagnosing Cushing's syndrome using facial image data, several challenges remain.
First, the limited sample size may result in insufficient statistical significance. 
Additionally, the preponderance of female samples over male samples could introduce gender bias, affecting the generalizability of the findings. 
Furthermore, since all data are sourced from a single center, the lack of multi-center validation may restrict the external validity of the results. 
Future research should aim to increase sample size, balance gender representation, and incorporate data from multiple centers to enhance the reliability and applicability of the findings.
Lastly, this study only utilized frontal views of volunteers, while profile views could also provide features relevant to the diagnosis of Cushing's syndrome. 
Additionally, 3D facial images can offer more spatial depth information. 
Therefore, future research could focus on the automatic diagnosis of Cushing's syndrome using multi-view fusion or 3D facial imaging.

\section{Conclusion}
This study provides a comprehensive analysis of the classification performance of DinoV2 and other pre-trained models in diagnosing Cushing's syndrome using facial images.
This study yields several key insights and establishes a foundation for future research on AI-based automated diagnosis of Cushing's syndrome.
Firstly, in experiments comparing convolutional neural networks, Transformer-based models and DINOv2 demonstrated superior performance, with the ViT model achieving the highest F1 score of 85.71
Moreover, our investigation into the impact of the freezing mechanism on DINOv2 performance revealed significant enhancements, indicating its strong generalization capability.
Finally, our analysis of potential gender bias in the models revealed that both DINOv2 and pre-trained models exhibit gender bias due to disparities in male and female patient data. 
This finding underscores the urgent need for more data, particularly relevant data from male patients.
In summary, this study offers valuable insights into the performance of advanced deep-learning models for the automated diagnosis of Cushing's syndrome. 
It highlights the critical role of training strategies in achieving optimal model performance. 
These findings are essential for the ongoing development and enhancement of machine learning models for the automated diagnosis of Cushing's syndrome.

\section{Acknowledgments}
This study was supported by the National key research \& development plan of China, major project of prevention and treatment for common diseases (2022YFC2505300, subproject: 2022YFC2505304)

\bibliographystyle{IEEEtran}
\bibliography{ref}
\end{document}